\documentclass[a4paper, 10pt, conference]{ieeeconf}
\pdfoutput=1 
\IEEEoverridecommandlockouts
\renewcommand{\baselinestretch}{0.93}

\usepackage{graphics} 
\usepackage{tikz}
\usetikzlibrary{shapes,arrows,positioning}
\usepackage{tkz-graph}
\usepackage{hyperref}  
\usepackage{url}
\usepackage{graphicx}
\usepackage[textsize=scriptsize,textwidth=1cm]{todonotes}
\usepackage{amsmath,amsfonts,amssymb,amsthm}

\usepackage{subcaption}
\usepackage{standalone}
\usepackage{siunitx}
\usepackage{pgfplots}
\pgfplotsset{compat=1.14}
\usepackage{makecell}
\usepackage{bbm}
\usepackage[noadjust]{cite}
\usepackage{graphicx}
\usepackage{multirow}
\usepackage[font=footnotesize,labelfont=bf]{caption}

\title{\LARGE \bf 
Explainability of Intelligent Transportation Systems using Knowledge Compilation: a Traffic Light Controller Case}

\author{Salom\'{o}n Wollenstein-Betech$^{1}$, Christian Muise$^2$, Christos G. Cassandras$^1$, \\ Ioannis Ch. Paschalidis$^1$, and Yasaman Khazaeni$^3$ 
\thanks{Supported in part by NSF under grants ECCS-1509084, DMS-1664644, CNS-1645681, and IIS-1914792, by AFOSR under grant FA9550-19-1-0158, by ARPA-E’s NEXTCAR program under grant DE-AR0000796, by the MathWorks, by the ONR under grant N00014-19-1-2571, and by the NIH under grant 1R01GM135930. }
\thanks{Research performed during an internship at the MIT-IBM Watson AI Lab.}
\thanks{$^{1}$Dept. of Electrical and Computer Engineering,
	Division of Systems Engineering, Boston University,
	Boston, MA, USA.
	{\tt\small \{salomonw, cgc, yannisp\} @bu.edu}}
\thanks{$^{2}$ School of Computing, Queen's University, Kingston, ON, Canada.  
{\tt\small christian.muise@queensu.ca}}
\thanks{$^{3}$ IBM Research AI, Cambridge, MA, USA.  
{\tt\small yasaman.khazaeni@us.ibm.com}}
}

\newcommand{\policy}{\mathcal{P}}
\newcommand{\states}{\mathcal{S}}
\newcommand{\data}{\mathcal{D}}
\newcommand{\state}[1]{s_{#1}}
\newcommand{\act}[1]{a_{#1}}
\newcommand{\sapair}[1]{\langle \state{#1}, \act{#1} \rangle}

\newcommand{\fluents}{\mathcal{F}}
\newcommand{\actions}{\mathcal{A}}

\newcommand{\hide}[1]{}

\theoremstyle{definition}

\newtheorem{definition}{Definition}

\begin{document}
    \maketitle
    \thispagestyle{empty}
    \pagestyle{empty}
    \begin{abstract}
Usage of automated controllers which make decisions on an environment are widespread and are often based on black-box models. We use Knowledge Compilation theory to bring explainability to the controller's decision given the state of the system. For this, we use simulated historical state-action data as input and build a compact and structured representation which relates states with actions. We implement this method in a Traffic Light Control scenario where the controller selects the light cycle by observing the presence (or absence) of vehicles in different regions of the incoming roads. \end{abstract}

    \section{Introduction}  \label{sec:intro}

Recent developments in computing power, algorithms and data handling have allowed for both accurate and complex automated decision-makers. These smart agents have been adopted widely in academia and industry to perform different tasks. With the same vigor as in other areas, these methods have been embraced to perform many tasks in the context of Smart Cities \cite{cassandras2016smart} and Intelligent Transportation Systems. Some examples include multi-agent traffic light controllers \cite{el2013multiagent} and re-balancing of Mobility-on-Demand systems \cite{wen2017rebalancing}. 

A typical model (e.g., deep reinforcement learning) uses high-dimensional inputs to provide powerful predictions or decisions to achieve the desired goals. Unfortunately, the trade-off between the complexity and the interpretation of the model often limits its adoption to many applications where stakeholders must trust and explain the decisions taken. 

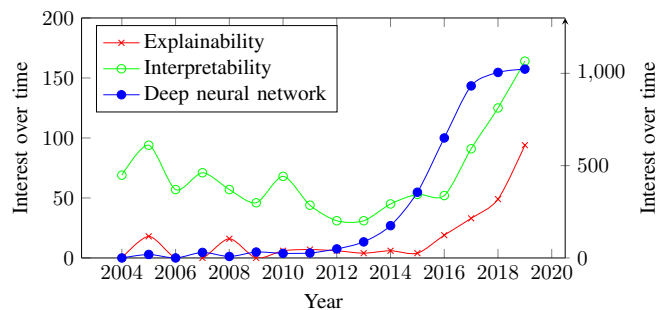
\begin{figure}[t]
    \centering
    \begin{tikzpicture}
\pgfplotsset{
    scale only axis,
}

\begin{axis}[
    axis y line*=left,
    ymin=0, ymax=200,
    xlabel= Year,
    ylabel=Interest over time,
    x tick label style={/pgf/number format/.cd,%
      scaled x ticks = false,
      set thousands separator={}},
    width=8cm,
    height=4cm
]
\addplot[smooth,mark=x,red]
  coordinates{
    (2004, 0)
    (2005, 18)
    (2006, 0)
    (2007, 0)
    (2008, 16)
    (2009, 0)
    (2010, 6)
    (2011, 7)
    (2012, 6)
    (2013, 4)
    (2014, 6)
    (2015, 4)
    (2016, 19)
    (2017, 33)
    (2018, 49)
    (2019, 94)
}; \label{plot_one}
\addplot[smooth,mark=o,green]
  coordinates{
    (2004, 69)
    (2005, 94)
    (2006, 57)
    (2007, 71)
    (2008, 57)
    (2009, 46)
    (2010, 68)
    (2011, 44)
    (2012, 31)
    (2013, 31)
    (2014, 45)
    (2015, 53)
    (2016, 52)
    (2017, 91)
    (2018, 125)
    (2019, 164)
}; \label{plot_two}
\end{axis}

\begin{axis}[
  axis y line=right,
  axis x line=none,
  ymin=0, ymax=1300,
  ylabel=Interest over time,
  width=8cm,
  height=4cm,
  legend pos= north west,
  legend cell align={left},
]

\addlegendimage{/pgfplots/refstyle=plot_one}\addlegendentry{Explainability}
\addlegendimage{/pgfplots/refstyle=plot_two}\addlegendentry{Interpretability}
\addplot[smooth,mark=*,blue]
  coordinates{
    (2004, 0)
    (2005, 19)
    (2006, 0)
    (2007, 30)
    (2008, 8)
    (2009, 32)
    (2010, 25)
    (2011, 27)
    (2012, 49)
    (2013, 87)
    (2014, 175)
    (2015, 356)
    (2016, 650)
    (2017, 932)
    (2018, 1005)
    (2019, 1023)
}; \addlegendentry{Deep neural network}

\end{axis}

\end{tikzpicture}
    \caption{Google Trend, read Explainability and Interpretability with left axis and Deep Neural Network with right axis}
    \label{fig:google-trend}
    \vspace{-5mm}
\end{figure}

The importance of explainability does not rely purely on justifying an agent's decisions. Understanding more about its controller logic provides clarity over unknown vulnerabilities of the controller. Also, it helps in identifying and correcting errors (debugging), thus enhancing the controller \cite{adadi2018peeking}.

With the success of non-interpretable methods (neural networks and boosting) the area of explainable AI (XAI) is now an exciting research topic, see Figure \ref{fig:google-trend}. The relevance of interpretability goes beyond the tech industry. For example, lawyers also benefit from understanding these models well when faced with claims about fairness of an agent's decision. As a response to this issue, the European Union has taken the lead by including a “right to an explanation” \cite{Goodman_Flaxman_2017} stated in articles 12-15 of the recently passed \emph{General Data Protection Regulation} (GDPR) law which pushes decision-maker architects to provide clear and concise statements of the logic involved.  

Within the transportation community, a tragic example is the \emph{Death of Elaine Herzberg} in Tempe, Arizona \cite{griggs_wakabayashi_2018}. This was the first recorded death provoked by a self-driving car. Many attribute the death to a mistake on the image classification method used by the vehicle. Nevertheless, we can't be sure of what exactly went wrong. This further encourages the interpretability of our intelligent transportation systems. 

In addition to these examples, we observe a growing trend in the relevance of  \emph{explainability} (or \emph{interpretability}\footnote{In this paper we use these terms interchangeably}) in google searches and in academia. To give one example, there was a 300\% growth in the number of papers presented at the \emph{Intelligent Transportation Systems Conference} in 2019 relative to the previous year, that contained these words. 

In this paper we present an interactive tool that helps bring interpretability to black-box controllers. We reason over the behavior of an agent by looking at historical data traces of state-action pairs without making assumptions on the system dynamics. Moreover, we assume the actions taken by the controller are fixed (i.e., a deterministic policy) but unknown. 

Related work has tried to solve similar problems on imitation or apprenticeship learning \cite{hanawal2018, abbeel2004apprenticeship}. The idea is to learn a policy from state-action samples, so that when the automated controller takes an action, it closely resembles the behavior of the original decision-maker. To achieve this goal, these methodologies require learning the dynamics of the system typically modeled as a Markov Decision Process (MDP). Our methodology differs, since it does not aim to learn the system dynamics (transition probabilities) nor how to control the agent. Rather, our goal is to provide useful and fast knowledge about the behavior of the agent on the environment. Our tool allows answering questions that other approaches lack. For example, we can ask: \emph{what is the probability of performing an action when a specific state variable is \texttt{True}?}.

We use the approach presented in \cite{muise-pair20}. The main idea is to use state-of-the-art knowledge compilation methods and to use disjunctive decomposable negation normal form (d-DNNF) as the key representation \cite{kcm} of the controller's logical theory.  This representation provides an interesting mix of compact representation, computationally efficient compilation, and expressive inference capabilities.

We bring this technology to the Transportation Community by analyzing a case study of the Traffic Light Control (TLC) problem. We chose to analyze this problem given the recent success of using Deep Reinforcement Learning (RL) to tackle the single-intersection TLC \cite{vidalideep} and multi-intersection TLC \cite{ prashanth2011reinforcement, shabestary2018deep, el2013multiagent}. 
However, despite Deep RL's great success on solving TLC, there is a limited amount of work attempting to explain the logic behind its trained controllers or predictors within the transportation domain. To the best of our knowledge, most attempts \cite{rizzo2019reinforcement, barredo2019lies} addressing the interpretability problem use Shapley Additive exPlanation (SHAP) methodology \cite{lundberg2017unified}. Our approach differs from SHAP in that we are, in a sense, constructing a hierarchical representation of interpretable insights rather than identifying the individual factors that contribute most to the output of the blackbox.

The rest of the paper is organized as follows. In Section \ref{sec:knowledge_compilation} we present background information on Knowledge Compilation and the d-DNNF language which serves as the basis for the interpretability model stated in Section \ref{sec:interpretability_model}. In Section \ref{sec:traffic_light_control} we present the TLC model, the RL approach used to train the smart controller, as well as the experiments performed.  In Section \ref{sec:results} we present some examples of explanations provided by the interpretability tool. We conclude in Section \ref{sec:conclusion} with a summary and with future directions.

\begin{figure*}[t]
\begin{subfigure}{0.32\textwidth}
\centering

\centering
\begin{tabular}{c c | c }
\hline
\makecell{\textbf{D}rive \\ mode  on}&\ \makecell{\textbf{K}ey \\ inside car} & Action \\ \hline
True & True & \textbf{dr}ive \\ 
False & True & \textbf{sw}itch\_to\_drive\_mode \\ 
False & False & \textbf{in}sert\_key \\ \hline
\end{tabular}
\caption{Data}
\label{fig:data}
\end{subfigure}
\begin{subfigure}{0.3\textwidth}
\centering
\begin{tikzpicture}
  [
  auto, every node/.style={draw,circle,minimum size=2em,inner sep=.1},node distance=.1cm, text=white, thick,
  and/.style={circle, draw=red!70, fill=red!70},
  or/.style={circle, draw=blue!70, fill=blue!70},
  activestate/.style={rectangle, draw=black!70, fill=black!70},
  nonactivestate/.style={rectangle, draw=gray!70, fill=gray!70},
  activeaction/.style={circle , draw=black!70, fill=black!70},
  nonactiveaction/.style={circle , draw=gray!70, fill=gray!70},
  ]

\node[or]{$\lor$}[level distance=10mm]{
      child[sibling distance=3.5cm] {node[and] {$\land$}
      child[sibling distance=1cm] {node[nonactivestate] {$\neg$D}}
      child[sibling distance=1cm] {node[nonactiveaction] {$\neg$dr}}
        child[sibling distance=1cm] {node[or] {$\lor$}
            child[sibling distance=2.8cm] {node[and] {$\land$}
                child[sibling distance=.8cm] {node[nonactivestate] {$\neg$K}}
                child[sibling distance=.8cm] {node[activeaction] {in}}
                child[sibling distance=.8cm] {node[nonactiveaction] {$\neg$sw}}
                }
            child[sibling distance=2.8cm] {node[and] {$\land$}
                child[sibling distance=.8cm] {node[activestate] {K}}
                child[sibling distance=.8cm] {node[nonactiveaction] {$\neg$in}}
                child[sibling distance=.8cm] {node[activeaction] {sw}}
                }
            }
       } 
      child[sibling distance=3.5cm] {node[and] {$\land$}
        child[sibling distance=.8cm] {node[activestate] {D}}
        child[sibling distance=.8cm] {node[activestate] {K}}
        child[sibling distance=.8cm] {node[nonactiveaction] {$\neg$in}}
        child[sibling distance=.8cm] {node[nonactiveaction] {$\neg$sw}}
        child[sibling distance=.8cm] {node[activeaction] {dr}}
      }
  }; 
\end{tikzpicture}
\caption{Compiled d-DNNF}
\label{fig:compiled}
\end{subfigure}
\begin{subfigure}{.3\textwidth}
\centering
\begin{tikzpicture}
  [
  auto, every node/.style={draw,circle,minimum size=2em,inner sep=.1},node distance=.1cm, text=white, thick,
  and/.style={circle, draw=red!70, fill=red!70},
  or/.style={circle, draw=blue!70, fill=blue!70},
  activestate/.style={rectangle, draw=black!70, fill=black!70},
  nonactivestate/.style={rectangle, draw=gray!70, fill=gray!70},
  activeaction/.style={circle , draw=black!70, fill=black!70},
  nonactiveaction/.style={circle , draw=gray!70, fill=gray!70},
  ]

\node[or]{$\lor$}[level distance=15mm]{
      child[sibling distance=3.5cm] {node[and] {$\land$}
        child[sibling distance=.8cm] {node[nonactivestate] {$\neg$D}}
        child[sibling distance=.8cm] {node[nonactiveaction] {$\neg$in}}
        child[sibling distance=.8cm] {node[activeaction] {sw}}
        child[sibling distance=.8cm] {node[nonactiveaction] {$\neg$dr}}
       } 
      child[sibling distance=3.5cm] {node[and] {$\land$}
        child[sibling distance=.8cm] {node[activestate] {D}}
        child[sibling distance=.8cm] {node[nonactiveaction] {$\neg$in}}
        child[sibling distance=.8cm] {node[nonactiveaction] {$\neg$sw}}
        child[sibling distance=.8cm] {node[activeaction] {dr}}
      }
  }; 
\end{tikzpicture}
\caption{Conditioned d-DNNF on K}
\label{fig:conditioned}
\end{subfigure}
\caption{Example compilation of agent behaviour. Let \emph{D}, \emph{K}, \emph{dr}, \emph{sw}, \emph{in} be \texttt{Drive\_mode\_on}, \hspace{.1cm}\texttt{Key\_inside\_car},  \hspace{.1cm}\texttt{drive}, \hspace{.1cm}\texttt{switch\_to\_drive\_mode},   and \texttt{insert\_key} respectively.
Figure \ref{fig:data} denotes the input $\data$ to the model, which we read every row as a conjunction. Figure \ref{fig:compiled} is the generated d-DNNF representation of the logical theory, which is compiled following the interpretability model depicted in Figure \ref{fig:approach}. To read the d-DNNF DAG, we suggest starting from top to bottom, and look at the left-most element on after a disjunction. This depicts the determinism property since the children of the \texttt{or} node are logically inconsistent. Lastly, Figure \ref{fig:conditioned} is the conditioned version of the d-DNNF when we set K to be active. One can observe that to get the conditioned tree we can prune the left branch of the lower \texttt{or} node in Figure \ref{fig:compiled}.}
\label{fig:image2}
\vspace{-6mm}
\end{figure*}
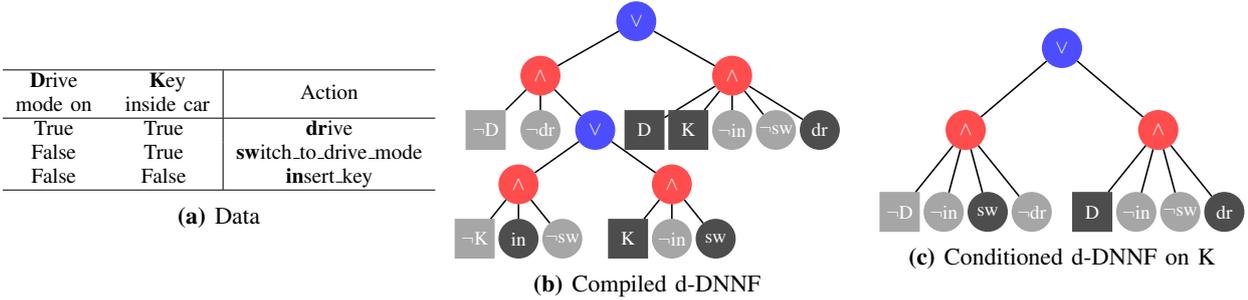
    
\section{Knowledge Compilation} \label{sec:knowledge_compilation}

\textbf{Background: }
The objective of Knowledge Compilation is to perform tractable operations of a complex logical theory. To achieve this, the technique builds structured representations of the logical theory in the form of a directed acyclic graph (DAG). The main idea is to compile off-line a complex and unorganized logical theory into a structured one, which is then used on-line to perform fast operations and reasoning. 

In the framework developed by \cite{muise-pair20}, the authors propose deterministic decomposable negation normal form (d-DNNF) as the target language to perform interpretation over a logical theory. The main characteristic of this language is that it allows to \emph{condition} and to \emph{count} the number of models in a propositional theory in polynomial time. These two operations are central in answering questions about an agent's behavior. Note that the off-line compilation to the d-DNNF language may be computationally expensive \cite{kcm}, however, this is just performed once. Let us now define properly the d-DNNF by introducing some of the languages used in Knowledge Compilation. 

\textbf{Languages: }
\subsubsection{Negation Normal Form (NNF)} \label{subsubsec:KC-lenguges-NNF}
Most languages in Knowledge Compilation are subsets of the NNF language \cite{kcm}. In this language, the only allowed Boolean operators are conjunctions ($\land$ , \texttt{and}) and disjunctions ($\lor$, \texttt{or}). The negation operator ($\neg$, \texttt{not}) is directly applied to the Boolean variables. In practice, these languages by are represented with a Directed Acyclic Graph (DAG). The graph describing a NNF theory is composed by leaves taking positive or negative boolean variables and inner nodes that are either conjunction or disjunction, see Figure \ref{fig:compiled} as an example. 

Formally, let $\Sigma$ be a  propositional theory (a DAG). Let $C$ be any node in $\Sigma$ and $\texttt{Vars}(C)$ be the set of variables appearing in the subgraph rooted at $C$.

\subsubsection{Disjunctive Normal Form (DNF)} \label{subsubsec:KC-lenguges-DNF}
This language is a subset of NNF and is formed by a disjunction of conjunctions (\texttt{or} of \texttt{and}'s), see Figure \ref{fig:conditioned} as an example. Its DAG is \emph{flat}, meaning that the distance from the root node to any leaf is $2$.  We represent the received data $\data$ using this language.  For each state-action pair $\langle \state{i}, \act{i} \rangle$ we build a clause $C_i$. Then, we take the disjunction over all of the $C_i$'s.

\subsubsection{Conjunctive Normal Form (CNF)} \label{subsubsec:KC-lenguges-CNF}
Similar to DNF, the CNF language is a conjunction of disjunctions (\texttt{and} of \texttt{or}'s) whose DAG is also \emph{flat}. The intent of using this intermediate language between our data (DNF) and target (d-DNNF) is twofold. (1) Ease the compilation CNF$\xrightarrow{}$d-DNNF by using ready-to-use compilers such as \textsc{Dsharp} \cite{dsharp-compiler}, c2d \cite{c2d-compiler} or D4 \cite{d4-compiler}, and (2) allow for encoding with different properties (see details on the different encoding flavors in \cite{muise-pair20}).

\subsubsection{Deterministic Decomposable Negation Normal Form (d-DNNF)} \label{subsubsec:KC-lenguges-d-DNNF}
This language is a subset of the NNF in which the properties of \emph{decomposability} and \emph{determinism} hold. The objective of having these two properties is to perform fast (polynomial time) operations of model counting and conditioning over the logical theory. The definition of these properties (as in \cite{kcm}) are:

\begin{definition}[Decomposability]
A NNF satisfies the decomposability property if for any conjunction $C$, the conjuncts of $C$ do not share any variable. In other words, if $C_1, ... , C_n$ are children of an \texttt{and} node $C$, then $\texttt{Vars}(C_i) \cap \texttt{Vars}(C_j) = \emptyset$ for $i \neq j$.
\end{definition}

\begin{definition}[Determinism]
A NNF satisfies the determinism property if for any disjunction $C$, every pair of disjuncts of $C$ are logically contradictory. That is, if $C_1, ... , C_n$ are children of an \texttt{or} node $C$, then $\forall i,j \in [1,...,n]$ where $i \neq j$, $C_i \land C_j = \texttt{False}$.
\end{definition}

\subsection{Logical operations in d-DNNF form} \label{subsec:KC-operations}
\subsubsection{Model Counting} \label{subsubsec:KC-operations-counting}
Without loss of generality, it is possible to count the number of models of a d-DNNF in polynomial time. Consider a logical theory $\Sigma$ and replace its \texttt{or} nodes by \emph{additions} and its \texttt{and} nodes by \emph{products}. Then, assign to each leaf the value of $1$. Note that an additional easy-to-comply property of \emph{smoothness} is needed to obtain a normalized count. 
\begin{definition}[Smoothness]
    A NNF satisfies the smoothness property if for each disjunction $C$, each disjunct of $C$ mentions the same variable. That is, $C_1, ... , C_n$ are children of an \texttt{or} node $C$, then $\texttt{Vars}(C_i) = \texttt{Vars}(C_j) \ \forall i,j \in [1,...,n]$ where $i \neq j$. 
\end{definition}
\subsubsection{Conditioning} \label{subsubsec:KC-operations-conditioning}
 Let $\Sigma$ be a d-DNNF and consider the problem of conditioning on variable $x$, i.e., we would like $\Sigma|x=\texttt{True}$. Then, we can replace $x$ by $\texttt{True}$ and $\neg x$ by $\texttt{False}$ and propagate this information throughout the DAG using standard logical rules. For example, if $\Sigma = (x \lor y) \land (\neg x \lor z)$,  then, $\Sigma|x = (\texttt{True} \lor y) \land (\texttt{False} \lor z)$ which simplifies to $z$.
Through the combination of conditioning and counting, the likelihood of a particular variable can be computed using
$
P(x=\texttt{True}) = \frac{count(\Sigma|x)}{count(\Sigma)}
$.

    \section{Interpretability Model} \label{sec:interpretability_model}

Recall that our goal is to reason over an agent's (traffic light) decisions and to answer questions about its underlying logic by analyzing samples of data. Given that we are using a model based on logical theories, the domain is restricted to discrete representations of states and actions. Nevertheless, one might consider using discretization to parse continuous to discrete domains. 

Let the set describing the state variables to be $\fluents$, the state-space (environment) be $\states \in  2^\fluents$ and a particular state to be $\state{} \in \states$. We use $\actions$ to represent the action space and $\act{} \in \actions$ to specify an action. Furthermore, let a state-action observation be a tuple $\sapair{}$ and our data be  $\data = \{\sapair{1},..., \sapair{m}\}$. The problem we face is then to succinctly represent the mapping $\policy: \states \rightarrow \actions$.
One intuitive way of solving the mapping problem would be to use a large table of state-actions. Unfortunately, this approach becomes quickly intractable due to the \emph{curse of dimensionality}. To overcome this computational burden, we use the model proposed by \cite{muise-pair20} based on Knowledge Compilation. 

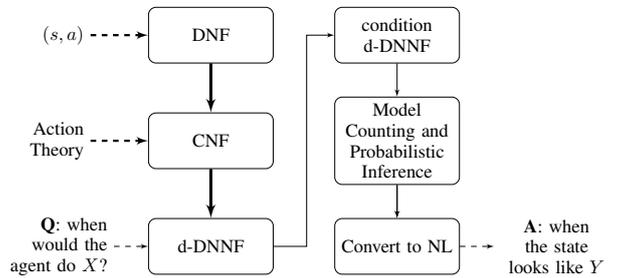
\begin{figure}
    \tikzstyle{block} = [rectangle, draw, 
    text width=6em, text centered, rounded corners, minimum height=3em]
\tikzstyle{line} = [draw, -latex']
\tikzstyle{input} = [text width=3.5em]
\tikzstyle{input_2} = [text width=6em]
   
\begin{tikzpicture}[node distance = 0cm, auto]
    \node [input,  align=right] (state-action) {$(s, a)$};
    \node [block, right of=state-action, node distance = 3cm] (DNF) {DNF};
    \node [block, below of=DNF,node distance = 2cm] (CNF) {CNF};
    \node [block, below of=CNF,node distance = 2cm] (d-DNNF) {d-DNNF};
    \node [block, right of=DNF, node distance = 3.5cm] (query-ddnnf) {condition d-DNNF};
    \node [block, below of=query-ddnnf, node distance=2cm] (model-counting) {Model Counting and Probabilistic Inference};
    \node [block, below of=model-counting, node distance=2cm] (convert-NL) {Convert to NL};

    \node [input, below of=state-action, node distance=2cm, align=right] (action-theory) {Action Theory}; 
    \node [input_2, below of=action-theory, node distance=2cm, align=right] (question) {\textbf{Q}: when would the agent do $X$?};
    \node [input_2, right of=convert-NL, node distance=3cm, align=center] (answer) {\textbf{A}: when the state looks like $Y$};
    \path [line, dashed, very thick] (state-action) -- (DNF);
    \path [line, ultra thick] (DNF) to (CNF);
    \path [line, ultra thick] (CNF) to (d-DNNF);
    \path [line] (d-DNNF)  -- ++(1.8,0) node(lowerright){} |- (query-ddnnf);
    \path [line] (query-ddnnf) to (model-counting);
    \path [line] (model-counting) to (convert-NL);
    \path [line] (model-counting) to (convert-NL);
    \path [line,dashed,very thick] (action-theory) to (CNF);
    \path [line,dashed] (question) to (d-DNNF);
    \path [line,dashed] (convert-NL) to (answer);
\end{tikzpicture}
    \caption{Flowchart of the logical model. Dotted arrows indicate inputs to the model whereas solid arrows indicate processing steps. Bold arrows indicate off-line (pre-processing) tasks.}
    \label{fig:approach}
    \vspace{-6mm}
\end{figure}

The processing framework in \cite{muise-pair20} uses the sampled data of state-action tuples as clauses in disjunctive normal form (DNF). Then, it compiles this theory to various flavors of conjunctive normal form (CNF), see \cite{muise-pair20} for different encodings and properties. Once the theory is in CNF form, it compiles it using any off-the-shelf compilers \cite{dsharp-compiler, d4-compiler} to produce a d-DNNF, which has the properties of \emph{decomposability} and \emph{determinism}. Then, easy operations of model counting and conditioning generate probabilistic inference responses on the behavior of the system. See Figure \ref{fig:approach} as a summary of the approach. As an example, consider the data received on Figure \ref{fig:data}, its corresponding d-DNNF representation on Figure \ref{fig:compiled}, and its conditioned response on Figure \ref{fig:conditioned}.

    \section{Traffic Light Control} \label{sec:traffic_light_control}

The Traffic Light Control (TLC) problem consists of dynamically adjusting a road's green and red light cycle to maximize the traffic flow trough an intersection. In the past, the problem has been tackled using estimates of traffic flows on each road for a specific time of the day \cite{robertson1969transyt, little1981maxband}. Today, with the ability to gather and communicate information in real time, traffic-responsive techniques are taking the lead on controlling these systems. See examples using Infinitesimal Perturbation Analysis \cite{fleck2015adaptive}, SCOOT \cite{hunt1981scoot} and many Reinforcement Learning variations for single \cite{vidalideep, prashanth2011reinforcement, shabestary2018deep}, and multi-intersection \cite{abdoos2011traffic, el2013multiagent} control.
The use of RL agents for traffic light control is motivated by the fact that agents can self-train. If these are trained correctly and for long enough samples, we can expect them to adapt to different situations including road accidents, weather and other variables. 
In particular, Deep RL differentiates itself by its ability to handle high-dimensional inputs. The RL agent learns to maximize a \emph{reward} function by observing the system state and by training a model that relate an output by using a complicated function of input variables.

\begin{figure}[t]
    \centering
    \includegraphics[width=\linewidth]{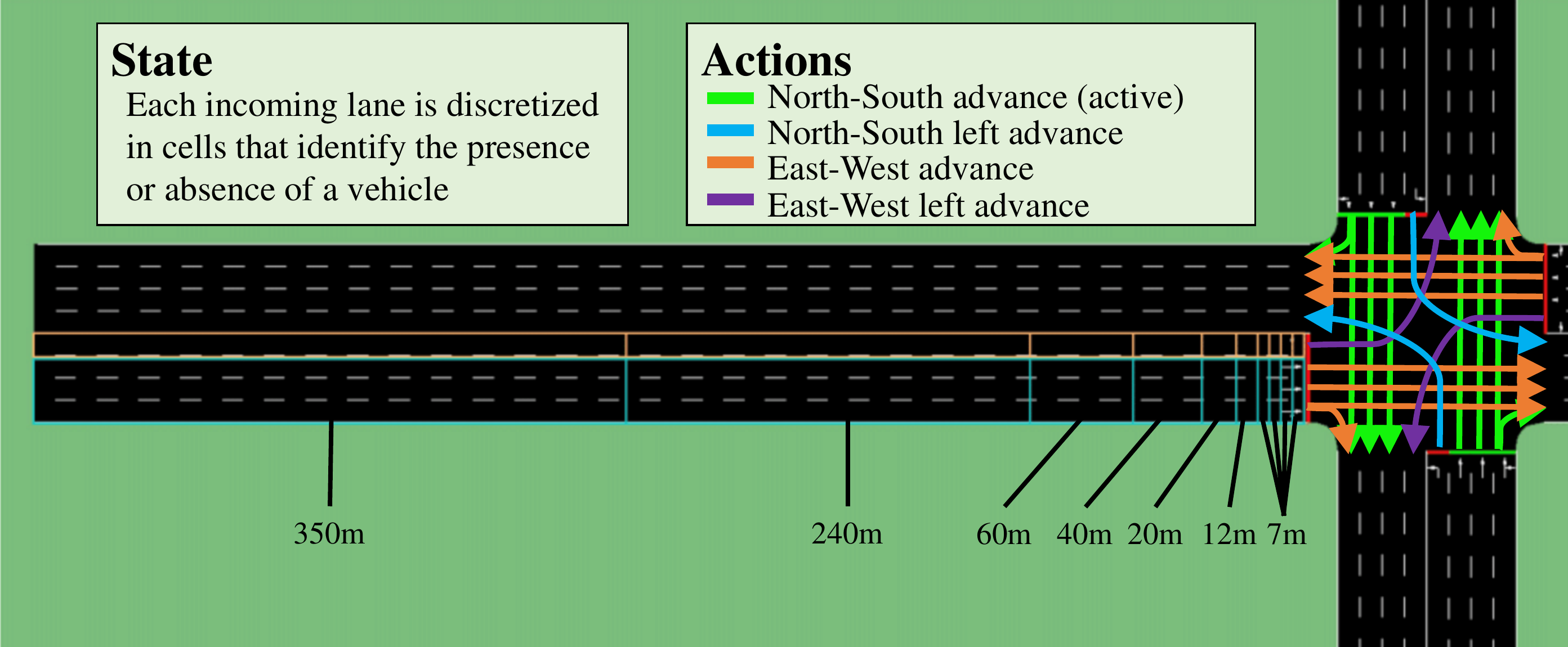}
    \caption{Traffic Light environment.}
    \label{fig:traffic-example}
    \vspace{-6mm}
\end{figure}

Given that our interpretability model needs both states and actions to be discrete, we would like to choose a TLC formulation that meets these requirements and facilitates the analysis. Hence, we consider a single intersection traffic light control scenario.
We build a simulation model using SUMO \cite{SUMO2018} consisting of a cross intersection with 4 incoming and outgoing lanes. Each incoming road to the intersection is set to be 750 meters long. We divide every road on the network into $n$ \emph{movements}. These \emph{movements} have predefined routes for all the cars flowing through a particular lane. In our case, let the left-most lane of every incoming edge be a movement (left turn) and the other lanes be another movement (keep straight), see Figure \ref{fig:traffic-example} as a reference. 
During the simulation, the agent (traffic light controller) samples the environment and receives a state $\state{t}$ and a reward $r_t$ at time $t$. According to this observation the agent chooses its next action $\act{t}$. At the same time the agent learns about the consequence of having chosen its previous action and updates its decision policy.

\textbf{State: }
Following the model in \cite{vidalideep} we let the state of the system be the collection of variables describing the presence or absence of vehicles on movement cells. We divide each incoming road into movements and we assume the total number of movements approaching the intersection is $m$. In our scenario $m=8$ as we divided each road in $2$ (vehicles turning left or keep  straight) and we have $4$ incoming roads. Then, we divide each of these movements $i=1,..,m$ into $b_i$ cells, which might differ in size (see Figure \ref{fig:traffic-example}). The choice of the length of a cell is not trivial. If cells are too long we have lower granularity, in contrast, when cells are too short it brings higher computational complexity which requires longer training times. 
The state is then $ \state{t} = \{ x_{ij}(t) \ | \ i=1,...,m ; \ j=1,..,b_i \}$ where the variable $x_{ij}(t)$ is equal to $1$ when there is at least one vehicle present on cell $j$ of movement $i$ at time $t$, and $0$ otherwise. Then the number of possible states in the TLC system is $|\states| = 2^{\sum_{i=1}^m b_i}$.

Note that our interpretability tool is not limited to this particular state-space. However, this representation meets our need for discrete state variables. Some other approaches include variables such as the relative velocity between vehicles \cite{gao2017adaptive}, current traffic light phase \cite{yau2017survey}, among others.

\textbf{Actions: }
We consider a single agent (the traffic light controller) which can choose between four possible actions. Each of these actions corresponds to a given configuration of red and green lights. The possible light phases are: North-South (NS), North-South left-most lane (NSL), East-West (EW), and East-West left-most lane (EWL). Hence $\actions = \{ \text{NS}, \text{NSL}, \text{EW}, \text{EWL} \}$. 
Once an action is selected, the traffic light will maintain this light phase for a fixed amount of time. We chose this length to be 10 seconds. Note, that the controller might choose the same action if for example, there is a lot of vehicles flowing on a particular direction. This does not impose an upper bound on the time a specific phase is active. In contrast, if the new selected action is different from the previous action, a 4 seconds yellow phase is initiated before starting the new phase. 
This delay allows drivers to anticipate and prepare before reaching the intersection. Additionally, it also benefits the model by preventing switching too often between light cycles. 

\textbf{Reward Function: }
In a RL setting, the reward function is the feedback the agent receives from the environment after performing an action. This feedback helps the agent improve its model of the environment in order to make better decisions in the future. 
In the setting of TLC, the objective is to maximize the traffic flow through the intersection over time. Hence, the reward is usually a measure of a vehicle's delay, queue lengths, waiting times or overall throughput which serve as proxies for traffic flow maximization. In this paper we use the reward function $r_t$ proposed in \cite{vidalideep}.

Let $w_\theta(i,t)$ be the cumulative time over which a vehicle $i$ has had speed smaller than $\theta$ up to time $t$. Then, assuming $n$ cars have arrived to the environment before time $t$, the total cumulative waiting time at $t$ is $ W_t(\theta) = \sum\limits_{i = 1}^n  w_\theta(i,t)$.
We define a reward function $r_t$ such that a positive value encourages an action and a negative value discourages it. Hence, a bad action can be represented by the increase in the cumulative waiting time when compared with the previous agent step (decision time). Let's assume that the agent would make decisions on the times defined by the sequence $\{t_1, t_2, ... \}$. Then the reward function at decision time $t_i$ is
$
    r_{t_i}(\theta) = W_{t_{i-1}}(\theta) - W_{t_{i}}(\theta)
$.

\textbf{Learning Process: }
\subsubsection{Model} \label{subsubsec:TLC-Learning-Model}
We use Deep $Q$-Learning as our learning algorithm. This technique combines $Q$-Learning and Deep Neural Networks. $Q$-learning is a basic form of Reinforcement Learning which uses $Q$-values to iteratively improve an agent's decision. These values are a learned metric of how good it is to take a particular action given a specific state and are formally expressed by
\begin{align*}
    Q(\state{t_i},\act{t_i}) \xleftarrow[]{} Q(\state{t_i},\act{t_i}) + \alpha(r_{t_{i+1}}(\theta) + \gamma \max_{\act{} \in \actions} Q(\state{t_{i+1}}, \act{}) \\- Q(\state{t_i}, \act{t_i}))
\end{align*}
where $\alpha$ is the learning rate, and $\gamma \in [0,1]$ is the discount factor used to leverage the importance of future rewards compared to the immediate one.

Often, computing all possible combinations of states and actions is intractable due to the curse of dimensionality which results in not having $Q$-values for some state-action pairs. To overcome this limitation, we estimate the $Q$-learning function using a deep neural network (DNN). We use the DNN architecture as in \cite{vidalideep}. This DNN is fully-connected and is composed by an input layer of $\sum_{i=1}^m b_i$ (the size of the state of the system) and 5 hidden layers of 400 neurons each using a rectified linear unit (ReLU) function. The output layer contains 4 neurons with a linear activation function representing the value of an action given a particular state. 

\subsubsection{Training} \label{subsubsec:TLC-Learning-Training}
We use \emph{Experience Replay} \cite{lin1992self} as our training method. This approach uses \emph{batch} learning instead of adjusting an agent's policy at every decision. In other words, rather than updating the policy at every step, the agent uses all gathered information to update only at pre-defined moments. We call every training cycle an \emph{episode} and define $E$ as the total number of \emph{episodes} used to train an RL agent.

In order to face the \emph{exploration-exploitation} trade-off, we use an $\epsilon$-greedy method with a linear exploration strategy. Let $e \in \{1,2,...,E\}$ indicate the current episode index, then the $\epsilon$ parameter at $e$ is expressed by $\epsilon_{e} = 1 - e/E$.
This method gives more weight to exploring at the beginning of the training phase, but as learning occurs, the agent exploits and reinforces its learned knowledge about the system. 

    \section{Experiments} \label{sec:experiments}

We perform experiments consisting on using the TLC model presented in Section \ref{sec:traffic_light_control} to train an efficient black-box agent to control a traffic light. Once the agent is trained, we compute multiple traces of state-action pairs which serve as the input data $\data$ for the interpretability model explained in Section \ref{sec:interpretability_model}. Then, we interact with the interpretabilty tool to reason over the underlying logic of the RL controller. 
To train the agent we use the simulation environment created in \cite{vidalideep}. We provide a diverse set of traffic flows aiming to learn real world scenarios. Moreover, we randomly increase and decrease the traffic generation on each of the incoming roads to create combinations of high and low traffic intensities on the roads. Each simulation (or episode) consists of $5,400$ seconds equivalent to $1.5$ hours. 

\begin{figure}[t]
    \centering
    \begin{tikzpicture}
\pgfplotsset{
    scale only axis,
}

\begin{axis}[
    ymin=-27000, ymax=0,
    xlabel= Episode,
    ylabel=Reward,
    width=8cm,
    height=4cm,
    legend pos= north west
]
\addplot[smooth,mark=x,red]
  coordinates{
(5.5, -26065.2)
(15.5, -21213.8)
(25.5, -16414.7)
(35.5, -14351.5)
(45.5, -12682.3)
(55.5, -11184.7)
(65.5, -7452.7)
(75.5, -6868.6)
(85.5, -6446.4)
(95.5, -5704.1)
}; \addlegendentry{Agent 1}
\addplot[smooth,mark=o,green]
  coordinates{
(5.5, -27071.7)
(15.5, -23285.1)
(25.5, -17865.6)
(35.5, -18273.3)
(45.5, -13565.9)
(55.5, -11968.5)
(65.5, -9804)
(75.5, -8366.6)
(85.5, -7321.7)
(95.5, -6533.3)
}; \addlegendentry{Agent 2}
\addplot[smooth,mark=+,blue]
  coordinates{
(5.5, -25343.6)
(15.5, -26338.2)
(25.5, -19154)
(35.5, -17687.8)
(45.5, -15612.8)
(55.5, -12666.5)
(65.5, -10833.6)
(75.5, -9156.9)
(85.5, -7982)
(95.5, -6876.6)
}; \addlegendentry{Agent 3}
\addplot[smooth,mark=+,magenta]
  coordinates{
(5.5, -23619.7)
(15.5, -21183.5)
(25.5, -17144.6)
(35.5, -15291.8)
(45.5, -12218.3)
(55.5, -10294.9)
(65.5, -9245.6)
(75.5, -7790.7)
(85.5, -7064.1)
(95.5, -6073.4)
}; \addlegendentry{Agent 5}
\addplot[smooth,mark=+]
  coordinates{
(5.5, -25197.6)
(15.5, -19792.3)
(25.5, -15780.4)
(35.5, -13160.2)
(45.5, -10581.8)
(55.5, -9304.5)
(65.5, -7710.7)
(75.5, -6768.3)
(85.5, -5737.4)
(95.5, -5213.9)
}; \addlegendentry{Agent 7}
\addplot[smooth,mark=+, color=orange]
  coordinates{
(5.5, -27656.4)
(15.5, -17557.7)
(25.5, -16116.4)
(35.5, -12182.9)
(45.5, -10535.5)
(55.5, -8822.3)
(65.5, -7764.7)
(75.5, -6830)
(85.5, -5614.3)
(95.5, -5456.2)

}; \addlegendentry{Agent 9}
\addplot[smooth,mark=+, color=yellow]
  coordinates{
(5.5, -23195)
(15.5, -19738.4)
(25.5, -15031.5)
(35.5, -12336.7)
(45.5, -11199.6)
(55.5, -9286.1)
(65.5, -8190.1)
(75.5, -7090)
(85.5, -6190.2)
(95.5, -5692.1)

}; \addlegendentry{Agent 10}
\end{axis}
\end{tikzpicture}
    \caption{Training different controllers. The agent number corresponds to the number of observable cells from the intersection on each movement.}
    \label{fig:training-traffic-example}
    \vspace{-6mm}
\end{figure}
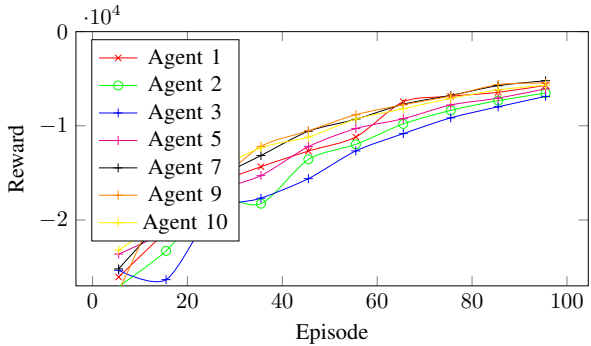

We consider the scenario with $8$ movements and divide each of these into $10$ cells (i.e., $n_i=10$ for all $i$) as in Figure \ref{fig:traffic-example}.
Based on this state-space representation we trained 7 different RL controllers aimed at assessing their performance and reasoning over their underlying behavior. The difference between these agents is their observation ability.
For each agent $l = 1,...,10$ we define their state-space to be the first $l$ cells of a movement $i$, formally,  $\state{t}(l) = \{ x_{ij}(t) \ | \ i=1,...,m ; \ j=1,..,l \} $. For example, agent $2$ will be trained on a state-space based on $x_{ij}(t)$ for $j=1,2$ exclusively. 
To assess the performance of each agent we learn over 100 episodes ($E=100$) and compare the reward trend while learning, see Figure \ref{fig:training-traffic-example}. An interesting observation about this training process is that Agent $2$ and $3$ have worse performance than Agent $1$ even though they observe everything $1$ does and more. We believe this behavior happens because the controller has difficulties differentiating between presence on a cell due to congestion or due to a passing vehicle. However, once the agent has more information further down the road, as Agent $10$ does, it is easier to differentiate the cause of the presence of a vehicle in a cell.

    \section{TLC interpretability results} \label{sec:results}

\begin{figure}[t]
    \centering
    \includegraphics[width=.98\linewidth]{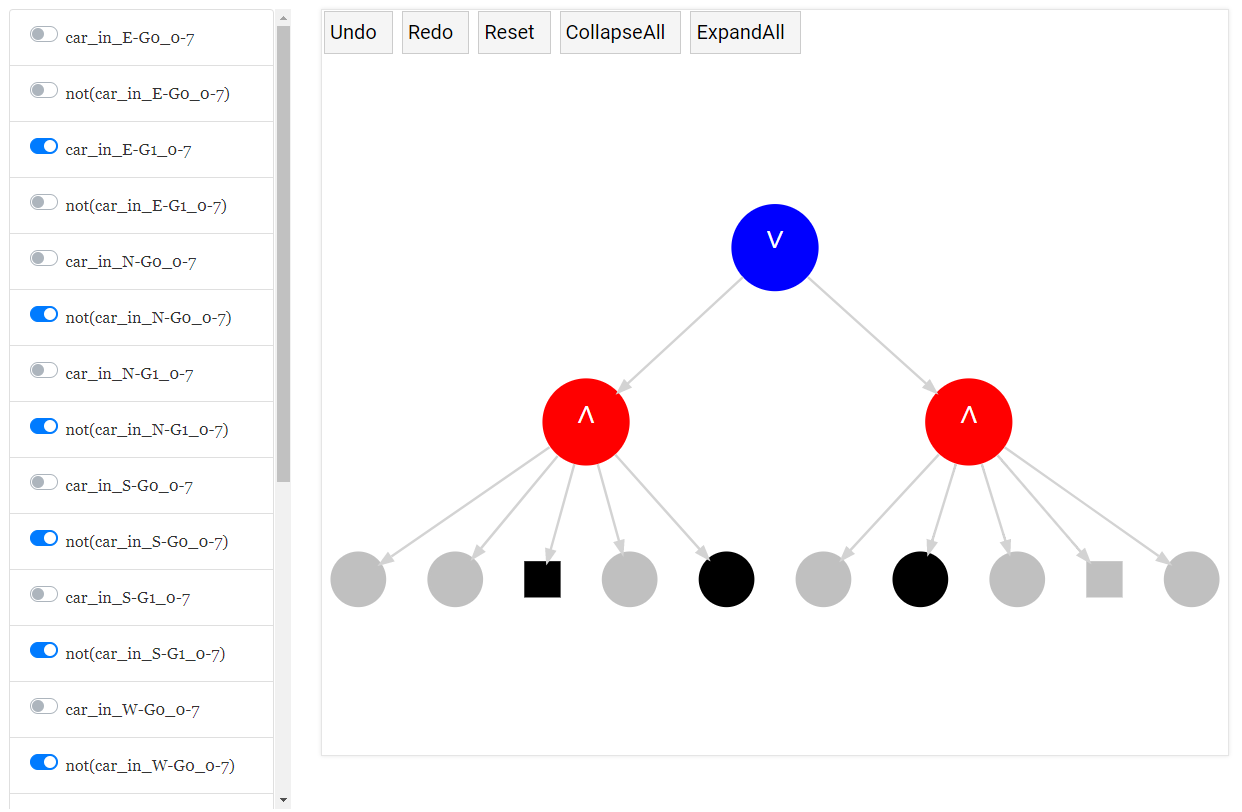}
    \caption{Screenshot of the controller logic of Agent 1 in the interactive tool. On the left hand, the user can toggle (on/off) to condition on a particular state variable or action. The controller is conditioned as stated in the first line of Table \ref{tab: action-likelihood}.}
    \label{fig:controler-logic-ddnnf}
    \vspace{-3mm}
\end{figure}

After training the controller, we ran $100$ simulations for each agent $l=1,...,10$ to create historical data $\data_l$. We used $\data_l$ as the input of the interpretability model to interact it with the  representation of the logical theory of the traffic light. 
The results obtained matched our intuition on how a TLC should operate. Take for example the first row in Table \ref{tab: action-likelihood}. In this case Agent 1 observes the first cell of each movement. We asked the interpretability model to give us the likelihood of an action when a vehicle is present on the first cell of the east-straight movement, unknown situation on the east-left movement, and no vehicles on any other movement. The result shows that the agent will choose either EW or EWL cycle with 50\% chance, given those condition. In Fig. \ref{fig:controler-logic-ddnnf}, we show the conditioned d-DNNF DAG for this query.  

\begin{table}[h]
    \centering
    \footnotesize
    \caption{All the queries assume unknown state variables unless specified in the table. Notation:  \textbf{R}-\textbf{G}\textbf{M}\_\textbf{N} stands for incoming \textbf{R}oad
    (N/S/E/W), \textbf{G}reen phase, \textbf{M}ovement (\textbf{0}=straight/\textbf{1}=left), and cell position $\textbf{N}$.}
    \begin{tabular}{lccccc}
        \hline
        \multirow{2}{*}{Query} & \multirow{2}{*}{Agent} & \multicolumn{4}{c}{Action Likelihood} \vspace{.5mm}  \\ \cline{3-6}
         &  & NS & NSL & EW & EWL \\ \hline
        \begin{tabular}[c]{@{}l@{}}Vehicle in E-G0\_0-7; \\ E-G1\_0-7 unknown;\\No vehicle in the rest\end{tabular} & 1 & 0.0\% & 0.0\% & 50.0\% & 50.0\% \\ \hline
        No conditioning & 7 & 33.6\% & 13.7\% & 37.3\% & 15.4\% \\ \hline
        \begin{tabular}[c]{@{}l@{}}Vehicle present in \\ E-G0\_0-7\end{tabular} & 7 & 21.4\% & 11.7\% & 58.2\% & 8.0\% \\ \hline
    \end{tabular}%
    \label{tab: action-likelihood}
    \vspace{-5mm}
\end{table}

More interestingly, on the second row of Table \ref{tab: action-likelihood}, we asked for the likelihood of an action without conditioning on anything. As expected, the time of EW vs. EWL (and NS vs. NSL) was not uniform.  This is because in the simulation, we consider higher vehicle flows going straight versus turning left (3 straight lanes  versus 1 turning left). Hence, if the RL controller tries to minimize the overall delay, we expect it to give preference to the straight trajectories over turning trajectories. From a debugging point of view we are satisfied with this result as it matches our intuition. 
The last row considers the presence of a car in the east-straight lane in position 1 for Agent 7, and all the other state variables are unspecified. Recall that Agent 7 observes for each road the first seven cells starting from the intersection.  We expect this query to give us information about the marginal gain that the action EW receives when the controller observes presence of a vehicle on the first cell (0-7 meters from the traffic light).  Intuitively, we expect the likelihood of action EW to be greater than the likelihood of other actions, which is exactly what we observe in the results. 

\begin{figure}[t]
    \centering
    \includegraphics[width=.98\linewidth]{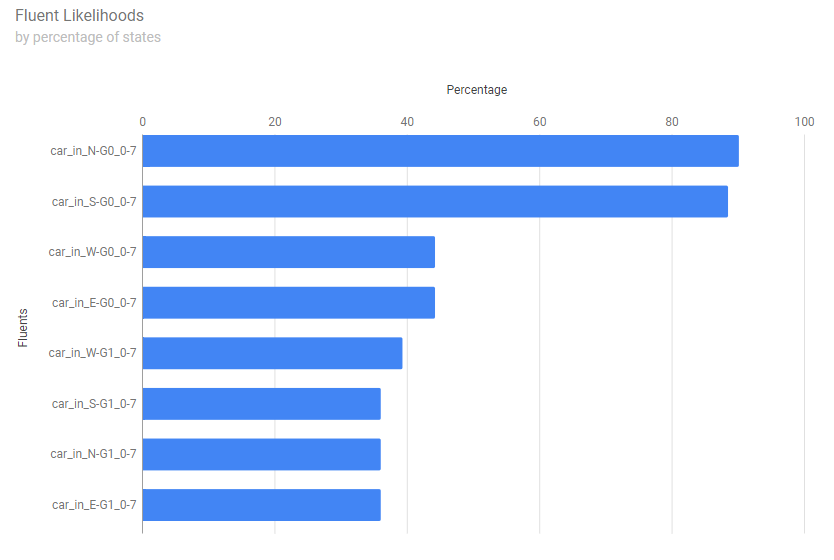}
    \caption{Environment likelihood of Agent 1 when it performs action NS. I.e., How does the state look like when Agent 1 performs action NS?.}
    \label{fig:state-likelihood}
    \vspace{-6mm}
\end{figure}

Another type of query we can ask our tool is to understand the environment conditional on an action being active. Take for example the question asked to Agent 1: \emph{What is the likelihood of a state variable when the controller decides on action NS?} The result of this query is shown in Figure \ref{fig:state-likelihood}. As we anticipate, we see that it is very likely to have vehicles in the north and south straight movements, whereas it is less likely to observe vehicles on the east, west or left movements.
In summary, the results obtained by the interpretability model match our intuition. The technique naturally lends itself to debugging black-box controllers, and ultimately aids in understanding the operation of the controller.

    \section{Conclusions} \label{sec:conclusion}
We use Knowledge Compilation techniques to reason over the behavior of black-box controllers. We present an example of this tool on a traffic light control setting where the controller uses a Deep Neural Network to decide its actions given a state. In this setting, the objective of the agent is to dynamically select the best light phase given a state composed by the presence (or absence) of vehicles in different cells of a road network. 
Once the black-box controller is trained, it is used to control the system. We then sample historical state-action pairs and use them as an input to the interpretability model. Then, we use the interactive tool which allows us to reason over the environment and the agent's decisions. It is worth pointing out that this general framework allows to reason over any type of decision-maker (including humans) and it is not reserved for a particular technique such as RL or any application such as TLC. 
For all the queries we performed, the interpretation given by the tool about the traffic light controller matches our intuition on the decisions we expect the agent must take given that information. Also, it provides a very nice platform for human-in-the-loop interaction with the system and we see its potential to be used for debugging purposes. 

\textbf{Future Work: }
We identify two interesting and important areas for future work. 
First, we would like to relax the assumption on policy determinism to allow for non-deterministic policies. This implementation will require weighted model counting which is a studied method within the knowledge compilation community.
Second, we would like to have this interpretability tool available for richer settings including continuous time action and state spaces. This will require to have a pre-processing phase in which the discretization of the domain occurs. This task is very complex as it requires an optimization on the thresholds of the state and action variables such that the interpretation of the controller is maximized.

    \bibliographystyle{IEEEtran}
    \begin{tiny}
        \bibliography{references}
    \end{tiny}
\end{document}